# Long-Term Prediction of Lane Change Maneuver Through a Multilayer Perceptron

Zhenyu Shou[1], Ziran Wang[2], *Member*, *IEEE*, Kyungtae Han[2], *Senior Member*, *IEEE*, Yongkang Liu[3], *Student Member*, *IEEE*, Prashant Tiwari[2], and Xuan Di[1], *Member*, *IEEE*

*Abstract* — **Behavior prediction plays an essential role in both autonomous driving systems and Advanced Driver Assistance Systems (ADAS), since it enhances vehicle's awareness of the imminent hazards in the surrounding environment. Many existing lane change prediction models take as input lateral or angle information and make short-term (< 5 seconds) maneuver predictions. In this study, we propose a longer-term (5~10 seconds) prediction model without any lateral or angle information. Three prediction models are introduced, including a logistic regression model, a multilayer perceptron (MLP) model, and a recurrent neural network (RNN) model, and their performances are compared by using the real-world NGSIM dataset. To properly label the trajectory data, this study proposes a new time-window labeling scheme by adding a time gap between positive and negative samples. Two approaches are also proposed to address the unstable prediction issue, where the aggressive approach propagates each positive prediction for certain seconds, while the conservative approach adopts a roll-window average to smooth the prediction. Evaluation results show that the developed prediction model is able to capture 75% of real lane change maneuvers with an average advanced prediction time of 8.05 seconds.**

## I. Introduction

Understanding other drivers' intention is one of the key issues faced by both autonomous driving systems and Advanced Driver Assistance Systems (ADAS). Drivers' maneuver typically consists of two main aspects, namely longitudinal maneuver (i.e., car-following or acceleration/deceleration) and lateral maneuver (i.e., lane change). As reported, lane change alone caused nearly 5% of crashes and 7% of crash fatalities in the road network [1], [2]. A reliable prediction model on drivers' maneuver intention can boost the safety of all participants in the road network, because drivers can take precaution to avoid any possible accidents based on the predicted maneuver intention of nearby drivers.

Maneuver prediction has been extensively studied in the literature [3]–[6]. Different studies use various features and classification or regression methods to predict maneuvers. Longitudinal maneuvers are continuous, and usually a regressor is employed to map some features to a value of acceleration or deceleration. For example, acceleration could be predicted by a feedforward neural network model [7], a long short-term memory (LSTM) neural network model [8], and a graph convolution model [9]. Lateral maneuvers are discrete, which typically consist of three actions (i.e., left lane change, right lane change, and lane keeping). Usually a classifier is formed to investigate the correlation between lateral maneuvers and some features. For example, lateral maneuver predictions were conducted by Bayesian networks [10], support vector machines (SVM) [11], [12], and neural networks [11], [13].

Compared with the prediction of longitudinal maneuvers, which is heavily correlated with the gap between the ego vehicle and the leading vehicle, the prediction of lateral maneuvers is more challenging. Although current studies on lane change prediction have achieved an acceptable prediction accuracy [10], [12]–[15], their prediction horizon is short-term (i.e., < 5 seconds), because lateral information and/or steering angle are used as features. To be precise, when the lateral position of a vehicle is approaching the lane mark or the steering angle of a vehicle points towards an adjacent lane, it is very likely that the vehicle will make a lane change soon.

To predict longer-term (5~10 seconds) lane change maneuvers, we propose a naturalistic neural network model without using lateral information or steering angle of a vehicle. In other words, only longitudinal positions, velocities, and accelerations are used in the feature set. Compared to the existing literature regarding this research topic, the main contributions of this study are as follows:

- Build a long-term (5~10 seconds) lane change intention prediction model without using lateral information or steering angle of a vehicle.
- Extend the time-window labeling scheme by adding a time gap between positive samples and negative samples to decorrelate the features of positive samples and that of negative samples.
- Propose an aggressive approach and a conservative approach to address the unstable prediction issue [9], and a comparison study is conducted between these two approaches.
- Evaluate the performance of the proposed model by NGSIM dataset which is collected in a real-world traffic environment.

The remainder of the paper is organized as follows. Section II introduces the related work. Section III presents three lane change prediction models. Section IV details the data and a cross validation of prediction models. Section V presents a run-time evaluation of the prediction model. Section VI concludes the paper with some insights regarding future work.

[1] Zhenyu Shou and Xuan Di are with the Department of Civil Engineering and Engineering Mechanics, Columbia University, New York, NY 10027, USA (e-mail: {zs2295, sharon.di}@columbia.edu)

[2] Ziran Wang, Kyungtae Han and Prashant Tiwari are with Toyota Motor North America, InfoTech Labs, Mountain View, CA 94043, USA (e-mails: {ziran.wang, kyungtae.han, prashant.tiwari}@toyota.com)

[3] Yongkang Liu is with the Electrical Department, University of Texas at Dallas, Richardson, TX 75080, USA (e-mail: yongkang.liu@utdallas.edu)

## II. RELATED WORK

The task of lane change maneuver prediction involves features and labels. Labels are lane change maneuvers, while features carry predictive information and are supposed to be correlated with labels. Features can be extracted from various sources, such as spatiotemporal knowledge from cameras, range information from Lidar/radar, vehicle dynamics (e.g. lateral and longitudinal accelerations) from inertial measurement unit (IMU), and steering wheel angle, gas/brake pedal positions from CAN-Bus. In this study, we propose a lane change prediction model without using the camera source, exploring the possibility to leverage limited information for accurate prediction. Furthermore, based on whether using lateral information or steering angle in features, we broadly categorize related studies into two groups.

The first group includes studies using lateral or angle information. In [14], all relevant features including vehicle features, lane features, and infrastructure features are considered and a feature ranking is performed by running single variable classifiers (Naïve Bayes or Hidden Markov Model). The authors found that among all relevant features, lateral velocity, lateral offset, and longitudinal velocity are the most import features that correlated with lane change maneuvers. In [10], a dynamic Bayesian network is proposed based on a rich feature set including direction of lateral velocity, yaw rate to the road tangent, etc. An advanced prediction time of 3.75 seconds is achieved. In [12], potential-based features are extracted by defining a dynamic characteristic potential field among vehicles. An SVM classifier is then adopted for the prediction task and can achieve a 98% F1 score and a 1.9 seconds advanced prediction time. In [16], lateral position and steering angle and their first derivatives are used as features in a prediction model consisting of an SVM classifier and a Bayesian filtering algorithm. The model is able to predict lane change maneuvers 1.3 seconds in advance on average. Similarly, in [13], lateral offset and steering angle of a vehicle are used in the feature set. The authors compared three machine learning techniques, namely an SVM, a recurrent neural network, and a feed-forward neural network for the prediction task and concluded that SVM performs the best based on their experiments.

The second group of studies do not include any lateral or angle information. In [11], an SVM classifier and an artificial neural network (ANN) classifier are adopted to predict the merging maneuver (i.e. left lane change) for vehicles merging from ramp to highway. Although lateral and angle information is not included in the feature set, the distance from the beginning of the ramp to the merging vehicle, which is strongly positively correlated with the merging probability, is used. To be precise, the longer the distance the more likely the vehicle performs merging maneuver. In [17], drivers' lane change intent is modeled through a Bayesian network with some features from the longitudinal positions and velocities of the ego vehicle and surrounding vehicles. The authors applied the prediction model to a two-lane scenario where drivers are asked to stay on the right lane unless overtaking a vehicle. In addition, drivers are categorized into three groups, namely aggressive, neutral, and conservative. The model can achieve a 78% accuracy and 4.5 seconds advanced prediction time.

## III. LANE CHANGE PREDICTION MODELS

This section aims to build a long-term lane change maneuver prediction framework. With trajectories of multiple vehicles, one can extract some features, denoted as a vector $\boldsymbol{x}_i$, and corresponding labels, denoted as a scalar $y_i$. The subscript $i \in \{1,2,\cdots,N\}$ signifies the $i^{th}$ data point, and in total there are $N$ data points. A classifier is then employed to uncover the correlation between $\boldsymbol{x}_i$ and $y_i$ and serves as the engine of the prediction framework. In this work, we compare three well-known classifiers, namely a logistic regression classifier, a multilayer perceptron (MLP) classifier, and a recurrent neural network (RNN) classifier [13], [18].

### A. Logistic Regression

Logistic regression is a commonly used classification tool to model the probability of a binary outcome (e.g., lane change or lane keeping in our context). Multinomial logistic regression is an extension of logistic regression to model the probability of an event with more than two outcomes (e.g., left lane change, lane keeping, and right lane change in our context). Multinomial logistic regression is formulated below, with the binary logistic regression as a special case of it.

Suppose there are $m$ possible outcomes of an event. Taking the first outcome as the reference, the log probability of the $k^{th}$ outcome is a linear combination of features, i.e.,

$$\log \frac{p_k}{p_1} = \boldsymbol{\beta}_k \cdot \boldsymbol{x}_i,$$

where $\boldsymbol{\beta}_k$ is a vector containing regression parameters. Taking exponential of both sides of the equation, one can derive the probability of the $k^{th}$ outcome under observation $\boldsymbol{x}_i$, i.e., $p_k = p_1 e^{\boldsymbol{\beta}_k \cdot \boldsymbol{x}_i}$. Considering that the summation of probabilities of all possible outcomes is one, i.e., $\sum_{k=1}^{m} p_k = 1$, one can derive $p_1 = \frac{1}{1+\sum_{k=2}^{m} e^{\boldsymbol{\beta}_k \cdot \boldsymbol{x}_i}}$ and $p_k = \frac{e^{\boldsymbol{\beta}_k \cdot \boldsymbol{x}_i}}{1+\sum_{k=2}^{m} e^{\boldsymbol{\beta}_k \cdot \boldsymbol{x}_i}}$ for $k \in \{2,3,\cdots,m\}$. With $N$ data points, i.e., $\boldsymbol{x}_i$ and $y_i$ for $i \in \{1,2,\cdots,N\}$, parameters $\boldsymbol{\beta}_k$ for $k \in \{2,3,\cdots,m\}$ are estimated by some optimization methods such as maximum likelihood.

### B. Multilayer Perceptron

MLPs are a type of feedforward neural networks, which are inspired by the brain mechanism from biology. An MLP consists of several layers, namely an input layer, some hidden layers, and an output layer. A single hidden layer MLP is schematically shown in Fig. 1. Each directed edge has an attached weight, which is a parameter to be learned.

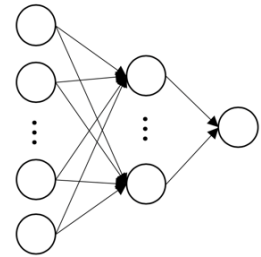

Fig. 1. A 3-layer MLP

To be concrete, we detail the definition for a 3-layer MLP used in this study with 4 hidden neurons in the hidden layer. Denote the weights and bias between the input layer and the hidden layer as a matrix $W_{12}$ and a vector $b_{12}$. Similarly, the weights and bias between the hidden layer and output layer are denoted as a matrix $W_{23}$ and a vector $b_{23}$. The input layer takes as input feature vector $\boldsymbol{x}_i$, and passes it through the directed edges. The hidden layer then receives $W_{12} \cdot \boldsymbol{x}_i + b_{12}$ and passes it through nonlinear activation function $\sigma_1$, and then to the directed edges. The output layer receives $W_{23} \cdot \sigma_1(W_{12} \cdot \boldsymbol{x}_i + b_{12}) + b_{23}$ and passes it through some nonlinear activation function $\sigma_2$. Thus, the output from this MLP given input $\boldsymbol{x}_i$ is $o_i = \sigma_2(W_{23} \cdot \sigma_1(W_{12} \cdot \boldsymbol{x}_i + b_{12}) + b_{23})$. With label $y_i$, a loss function can then be defined, e.g., $L = \sqrt{\frac{1}{N}\sum_{i=1}^{N}(y_i - o_i)^2}$. Backward propagation can then be used to solve for the weights and bias, namely $W_{12}$, $W_{23}$, $b_{12}$, and $b_{23}$.

### C. Recurrent Neural Network

To capture the temporal characteristics of lane change maneuvers, an RNN classifier is also adopted. RNNs are capable of processing a sequence by a hidden state which serves as the memory part of the neural network. Fig. 2 presents an RNN and the corresponding unrolled version.

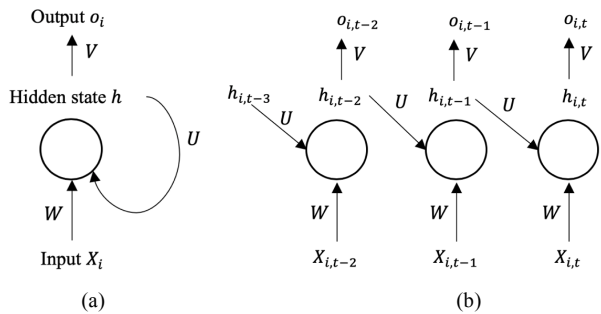

Fig. 2. Illustrations of (a) an RNN cell, and (b) the corresponding unrolled representation

Note that RNN takes as input a sequence of features, the input $X_i$ is now a matrix with each column representing features at a given timestep $t$. At each timestep $t$, RNN takes as input the weighted average of the feature set $X_{i,t}$ and a hidden state from the previous timestep, i.e., $W \cdot X_{i,t} + U \cdot h_{t-1}$, passes it through a nonlinear activation function $\sigma$, and generates a new hidden state $h_t$. $W$ and $U$ are two parametric matrices to be determined. After collecting all hidden states at different time steps, one can simply use another parametric matrix $V$ to convert hidden states to outputs. Depending on specific applications, final output from RNN can be a sequence of variable length or a scalar by selecting a certain part of outputs. In our context, we aim to predict a lane change maneuver based on the input feature set at timestep $t$, i.e., $X_i$. Thus, only the last element in outputs is kept as the prediction, i.e., $o_i = o_{i,t}$. With the corresponding label $y_i$, a loss function can then be defined, e.g., $L = \sqrt{\frac{1}{N}\sum_{i=1}^{N}(y_i - o_i)^2}$. Similar to MLP, some backward propagation algorithm can be used to solve for the parametric matrices, namely $W$, $U$, and $V$.

## IV. DATA AND EXPERIMENTS

In this section, we aim to evaluate the performance of aforementioned models using a real-world dataset. The processing, feature extraction and labeling of the dataset are introduced first. Then, we analyze the performances of different models and associated labeling schemes.

### A. NGSIM Data

*1) Data introduction:* The dataset used in this study is the reconstructed NGSIM I-80 vehicle trajectory data during an afternoon peak period 4 - 4:15 PM [19], [20]. The original NGSIM I-80 data from U.S. Federal Highway Administration (FHWA) was collected at a frequency of 10 Hz on Interstate 80 in Emeryville, CA, on April 13, 2005. The study area was a 500 m long highway segment with six main lanes (including a high-occupancy (HOV) lane and an on-ramp). Seven cameras were installed to record trajectories of all vehicles passing through the study area. Some computer vision techniques were used to process the video data and extract useful information such as position, velocity, and acceleration of all vehicles. The extracted information, however, inevitably contains noise. Thus, the reconstructed data that was filtered by some techniques is used in this study. Interested readers are referred to [19], [20] for details of the reconstructing process.

A short segment of data sample is listed in Table I, where $id$ is the vehicle identification, $frame$ is the current time frame, $l$ is the lane id, $pos$ is the longitudinal position of the vehicle, $v$ is the velocity, $a$ is the acceleration, $d_h$ is the headway, $c$ is the vehicle class, $n_f$ is the follower, and $n_l$ is the leader. For example, the first row shows that at frame 244, vehicle 4 is in lane 5 at a longitudinal position of 71.26 and is moving at a velocity of 7.40 m/s and an acceleration of 0.005 m/s$^2$. Vehicle 4 has a headway of 4.08 m. There are one follower (vehicle 27) and one leader (vehicle 21).

Fig. 3 presents a snapshot of vehicles in the study area at frame 2000. There are two type of vehicles, namely passenger vehicles shown as squares and heavy-duty trucks shown as circles. The blue and red color indicate that the vehicle is moving forward and performing left lane change, respectively. Lane 1 is the HOV lane, and lane 7 is the on-ramp. From the visualization, one can see that except the HOV lane and the on-ramp, vehicles are quite densely distributed.

TABLE I
A SHORT SEGMENT OF NGSIM RECONSTRUCTED DATA SAMPLE

| id | frame | l | pos | v | a | $d_h$ | c | $n_f$ | $n_l$ |
|----|-------|---|-------|------|-------|------|---|----|----|
| 4 | 244 | 5 | 71.26 | 7.40 | 0.005 | 4.08 | 2 | 27 | 21 |
| 4 | 245 | 5 | 72.00 | 7.40 | 0.011 | 4.08 | 2 | 27 | 21 |

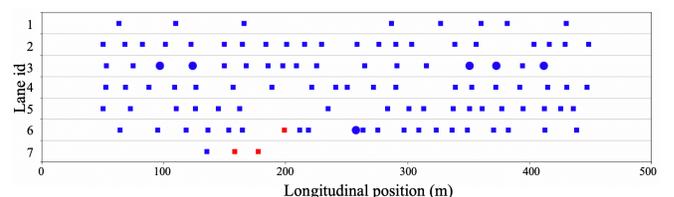

Fig. 3. A snapshot of vehicles on I-80 at frame 2000

*2) Data preprocessing:* Since vehicles on the on-ramp lane (lane id = 7) are forced to merge when the lane ends, we remove the rows with lane id 7 for precise evaluation. During the 15-minute study period, there were 652 left lane change maneuvers and 136 right lane change maneuvers. Considering the data insufficiency of right lane change, we focus on the prediction of left lane change in this study. Therefore, the rows with lane id 1 are further removed, because there is no left lane available for vehicles on lane 1.

For an ego vehicle performing left lane change, there are various driving scenarios. Fig. 4 presents one driving scenario where there are one leading vehicle on the current lane (i.e., vehicle 1), one leading vehicle (i.e., vehicle 2) and one following vehicle (i.e., vehicle 3) on the left lane, from the perspective of the ego vehicle (i.e., vehicle 0). The existence of vehicles 1, 2, and 3 results in 8 (i.e., $2^3$) different driving scenarios: a) all vehicles 1, 2, and 3 exist, b) vehicles 1 and 2 exist, c) vehicles 1 and 3 exist, d) vehicles 2 and 3 exist, e) vehicle 1 exists, f) vehicle 2 exists, g) vehicle 3 exists, and h) no vehicle exists. Based on data, the driving scenario shown in Fig. 4 consists of around 88% of all cases. Thus, we focus on the prediction of left lane change for this specific driving scenario.

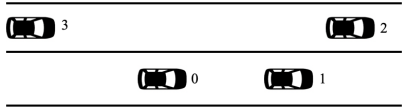

Fig. 4. The driving scenario of interest

*3) Feature extraction and labeling:* For the driving scenario of interest shown in Fig. 4, several features such as gap, velocity difference, and acceleration difference are extracted from the data. Feature selection is then performed by employing the feature selection module in scikit-learn [21], and the selected features are listed in Table II.

TABLE II
EXTRACTED FEATURES FOR LANE CHANGE PREDICTION

| Feature | Explanation |
|---|---|
| $d_{01}$ | Longitudinal distance between vehicle 0 and vehicle 1 |
| $d_{02}$ | Longitudinal distance between vehicle 0 and vehicle 2 |
| $d_{03}$ | Longitudinal distance between vehicle 0 and vehicle 3 |
| $v_{01}$ | Velocity difference between vehicle 0 and vehicle 1 |
| $v_{02}$ | Velocity difference between vehicle 0 and vehicle 2 |
| $v_{03}$ | Velocity difference between vehicle 0 and vehicle 3 |
| $l_0$ | Lane id of vehicle 0 |

With features at each frame extracted, now our task is to label each feature before we can perform any prediction. Although extracting the lane change maneuver from the data is straightforward, there is no consensus on how long it takes for a driver to prepare before he/she actually makes the lane change. In [17], a gaze-based labeling and a time-window labeling method were compared. The authors found that the former is more effective in their two-lane experiment. Nonetheless, the gaze-based labeling method requires some monitoring of drivers' gaze behavior, which is typically not accessible in a real-world driving scenario. Therefore, a time-window labeling method is adopted in this study.

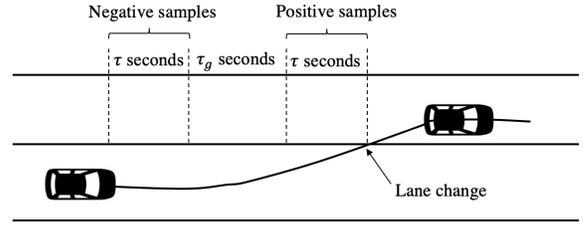

Fig. 5. A time-window labeling method

Fig. 5 presents the time-window labeling method used in this study. Data within a time-window of size $\tau$ seconds with the right end at the lane change point is labeled as positive samples. Moving the time-window $\tau + \tau_g$ seconds backwards, one can obtain negative samples.

Different from directly labeling $\tau$ seconds before the positive samples as negative samples [17] (i.e., $\tau_g = 0$), we now assign a gap of $\tau_g$ between positive and negative samples. The rationale is as follows. If $\tau_g = 0$, the last several frames in negative samples and the first several frames in positive samples share very similar features, because usually traffic condition does not change too much during a very short period of time. They, however, have different labels. In other words, similar features are mapped to different labels, resulting in difficulties in training the prediction model. Adding a gap of $\tau_g$ between positive and negative samples, the features of negative samples and that of positive samples are more uncorrelated. Note that the same time-window size is used for positive and negative samples for the purpose of balancing the number of positive and negative samples, which is critical for training the prediction model.

4) *Training and testing data:* To test the performance of prediction models on unseen data, we first randomly split the data into a training set and a testing set by a ratio of 4:1. To select the best model and the best labeling scheme (i.e., optimal $\tau_g$), we perform a $k$-fold cross validation on the training set. The procedure goes as follows. We first randomly shuffle the training data and split it into $k$ folds. For each fold, we treat it as the held-out validation data and use the remaining $k-1$ folds to train the prediction model, and then test the performance of the trained model on the held-out validation data. Considering that the total number of left lane change in the training set is 521 (i.e., $652 \times 0.8$), we choose $k$ as 5 in this study to ensure enough positive samples in the validation data. With the 5-fold cross validation, we can choose the best model and best labeling scheme and employ them in a run-time evaluation on the unused testing data in next section.

### B. Performance of Prediction Models

*1) Performance metrics:* Before comparing the performance of prediction models, we define several performance metrics as follows [9].

- True positive (TP): number of correct left lane change predictions.

- False positive (FP): number of wrong left lane change predictions.
- True negative (TN): number of correct lane keeping predictions.
- False negative (FN): number of wrong lane keeping predictions.
- True positive rate: $TPR = \frac{TP}{TP+FN}$.
- False positive rate: $FPR = \frac{FP}{FP+TN}$.
- Precision: $PRE = \frac{TP}{TP+FP}$.
- F1 score: $F1 = \frac{2 \times PRE \times TPR}{PRE+TPR}$.
- Accuracy: $Accuracy = \frac{TP+TN}{TP+FP+TN+FN}$.

2) *Results:* In addition to the comparison among different prediction models, we also compare the performance of different labeling schemes (i.e., different values of $\tau_g$). To be specific, we choose a fixed time-window size $\tau$ as 5 seconds [16] and vary the gap between positive samples and negative samples, i.e., $\tau_g$ from 0 to 15 seconds with an increment of 5 seconds. In other words, we test four labeling schemes (LS), namely LS1 (i.e., labeling scheme 1) with $\tau_g = 0$ seconds, LS2 with $\tau_g = 5$ seconds, LS3 with $\tau_g = 10$ seconds, and LS4 with $\tau_g = 15$ seconds.

Fig. 6 presents results from logistic regression. Using different labeling schemes, the performance varies. LS4 outperforms other labeling schemes in terms of both the F1 score and accuracy. The overall F1 score and accuracy achieved by using LS4 are 0.64 and 0.62, respectively.

Fig. 7 presents results from MLP. LS3 and LS4 outperform other labeling schemes in terms of both the F1 score and accuracy. Overall, LS3 and LS4 achieve similar F1 score and accuracy, The F1 score and accuracy achieved by using LS3 or LS4 are 0.72 and 0.71, respectively, which are significantly better than that achieved by logistic regression.

Fig. 8 presents results from RNN. LS4 outperforms other labeling schemes in terms of both the F1 score and accuracy. The overall F1 score and accuracy achieved by using LS4 are 0.59 and 0.64, respectively, which are lower than that achieved by MLP.

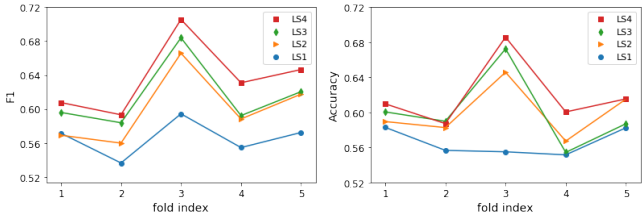
Fig. 6. Results from logistic regression: F1 score (left) and accuracy (right)

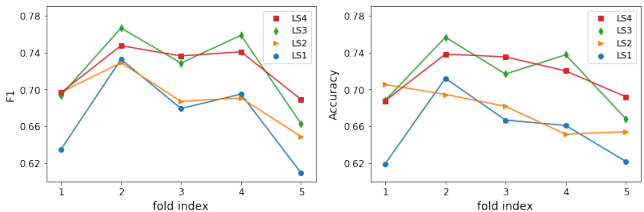
Fig. 7. Results from MLP: F1 score (left) and accuracy (right)

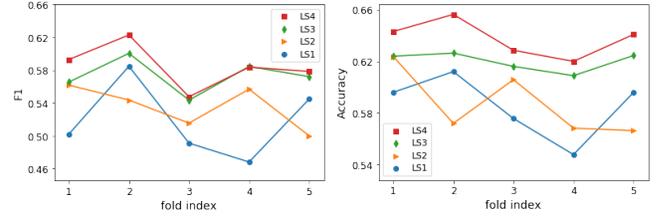
Fig. 8. Results from RNN: F1 score (left) and accuracy (right)

The overall outperformance of LS4 and LS3 over other labeling schemes are as expected. As previously mentioned, adding a proper gap between positive samples and negative samples reduces the correlation between the features of positive samples and that of negative samples, meaning that it is now easier for learning algorithms (i.e., logistic regression, MLP, or RNN) to effectively distinguish the features labeled positive samples from that labeled negative samples.

The outperformance of MLP over logistic regression and RNN is explained as follows. The correlation between features and the corresponding label in the prediction of lane change is complex and nonlinear. MLP is able to capture the nonlinearity which is missed by logistic regression. RNN captures the time series characteristics, however, it requires a large amount of training data. In the context of lane change prediction using NGSIM data, the number of lane change maneuvers (i.e., 652 left lane change maneuvers and 136 right lane change maneuvers) is not sufficient to reasonably train an RNN prediction model.

## V. RUN-TIME EVALUATION

In addition to the five-fold cross validation on the training set, a run-time evaluation on the testing set further tests the effectiveness of lane change prediction models in near real-time validation. The procedure of a run-time evaluation is as follows. With trajectory data of all 1,630 vehicles in the training set, we use LS4 (i.e., the best labeling scheme in the cross validation) to label the training data. The testing data of 407 vehicles in the testing set is not labeled in the run-time evaluation. After training the MLP prediction model (i.e., the best prediction model in the cross validation) using the labeled training data, we use the model to predict lane change maneuvers at every frame for each vehicle in the testing set.

### A. Modified Predictions Methodology

Table III presents a prediction sample of vehicle ID 447, where 0 and 1 mean negative and positive predictions, respectively. Note that in this case study, the model makes one lane change prediction per second. The vehicle made a real left lane change at frame 1910. Our prediction model predicts the vehicle to perform lane change at frames 1870 and 1910. As one can see, this prediction is unstable in the sense that the positive predictions are not consecutive. The purpose of predictions is to help drivers drive safely and at ease. Predictions in Table III do advise the vehicle driver to act cautiously at frame 1870. However, it suggests no precaution during frames 1880 to 1900, which is not helpful but confusing from the perspective of the vehicle driver. To overcome this unstable prediction issue, we propose two approaches, namely an aggressive approach and a conservative approach.

TABLE III
PREDICTION SAMPLE

| id | frame | LC (real) | LC (predicted) | Aggressive-Approach Prediction | Conservative-Approach Prediction |
|---|---|---|---|---|---|
| 447 | 1840 | 0 | 0 | 0 | 0 |
| 447 | 1850 | 0 | 0 | 0 | 0 |
| 447 | 1860 | 0 | 0 | 0 | 0 |
| 447 | 1870 | 0 | 1 | 1 | 0 |
| 447 | 1880 | 0 | 0 | 1 | 0 |
| 447 | 1890 | 0 | 0 | 1 | 0 |
| 447 | 1900 | 0 | 0 | 1 | 0 |
| 447 | 1910 | 1 | 1 | 1 | 0 |

The aggressive approach is presented in *Algorithm 1*. It propagates a positive prediction for another $\tau_a$ seconds. In other words, each positive prediction lasts for $\tau_a + 1$ seconds. For example, the fifth column in Table III presents the application of the aggressive approach to the prediction in the fourth column with $\tau_a = 3$. The conservative approach is presented in *Algorithm 2*. It smooths out the prediction through an average fashion. At each frame, we look at both the current prediction and the predictions for the previous $\tau_c$ seconds and take the average of these $\tau_c + 1$ values. If the average value is greater than a threshold value, then the actual prediction at this frame is positive. Otherwise, it is a negative prediction. For example, predictions are 1 at frame 1870 and 0s at previous $\tau_c = 3$ frames. The average is thus 1/4 which is below the threshold 0.5 used in this study. We therefore modify the prediction at frame 1870 from 1 to 0.

Although the aggressive or the conservative approach is able to yield more consecutive predictions, there are limitations in both approaches. For the aggressive approach, it adds more positive predictions, leading to a higher false positive rate (i.e., increased false alarm); while for the conservative approach, it suppresses some positive predictions, resulting in a lower true positive rate (i.e., decreased prediction accuracy). Thus, there exists a tradeoff between these two approaches, and both parameters, namely $\tau_a$ and $\tau_c$ need to be carefully chosen depending on specific applications.

*Algorithm 1*: **Aggressive approach**
**Input**: Lane change predictions $lc_t$ from the model at each frame $t \in \{1,2,\cdots,T\}$, parameter $\tau_a$.
**Output**: Modified lane change predictions $\widehat{lc}_t$
01: Initialize $\widehat{lc}_t$ as zeros
02: **for** $t \in \{1,2,\cdots,T\}$ **do**
03:   **if** $lc_t = 1$ **then**
04:     **for** $\tau \in \{0,1,\cdots,\tau_a\}$ **do**
05:       **if** $t + \tau \leq T$ **then**
06:         $\widehat{lc}_{t+\tau} = 1$
07:       **end if**
08:     **end for**
09:   **end if**
10: **end for**
11: **return** $\widehat{lc}_t$

*Algorithm 2*: **Conservative approach**
**Input**: Lane change predictions $lc_t$ from the model at each frame $t \in \{1,2,\cdots,T\}$, parameter $\tau_c$, and threshold value $thres$
**Output**: Modified lane change predictions $\widehat{lc}_t$
01: Initialize $\widehat{lc}_t$ as zeros
02: **for** $t \in \{\tau_c, \tau_c + 1, \cdots, T\}$ **do**
03:   $s = 0$
04:   **for** $\tau \in \{0,1,\cdots,\tau_c\}$ **do**
05:     $s \mathrel{+}= lc_{t-\tau}$
06:   **end for**
07:   $avg = s/(\tau_c + 1)$
08:   **if** $avg > thres$ **then**
09:     $\widehat{lc}_t = 1$
10:   **end if**
11: **end for**
12: **return** $\widehat{lc}_t$

### B. Strict Testing Criteria

In the run-time evaluation, two most important performance metrics are the prediction accuracy and the advanced prediction time. Different from the accuracy used in the previous five-fold cross validation, the prediction accuracy here is defined as the ratio of number of correct predictions to the number of total real lane change maneuvers. The advanced prediction time is defined as $t_{lc} - t_p$ where $t_{lc}$ is the time of a real lane change and $t_p$ is time of the earliest positive prediction that satisfies the following condition: predictions within the range of $[t_p, t_{lc}]$ are all positive.

For a real lane change maneuver, the definition of a correct prediction is the key in the evaluation. Some loose criteria are used in the literature. For example, in [17], as long as the vehicle makes a real lane change within 8 seconds of a positive prediction, then it is a correct prediction. Using this criterion, the prediction presented in the fourth column of Table III is classified as a correct prediction, which is obviously not the real case. Thus, in this study, we define a strict criterion as follows. For a real lane change maneuver at frame $t$, if and only if all predictions from frame $t - \tau_p$ to frame $t$ are positive, we call this a correctly predicted lane change; otherwise the real lane change maneuver is not correctly predicted. In this study, we take $\tau_p = 3$. For example, the fifth column in Table III is a correct lane change prediction while the fourth column is not.

### C. Results

Table IV lists the prediction results for the run-time evaluation for three approaches, namely plain predictions from the MLP model, predictions corrected by the aggressive approach, and the predictions smoothed by the conservative approach. Plain MLP model achieves a 62% prediction accuracy with an average advanced prediction time of 6.35 seconds. Compared with the plain MLP model, the predictions corrected by the aggressive approach substantially increases the prediction accuracy to 75% with a much higher average advanced prediction time of 8.05 seconds, although at an increased cost in FPR from 0.32 to 0.46 (i.e., increased false alarm). The conservative approach seems to be less effective in this case because false alarm is only slightly decreased (i.e., from 0.32 to 0.31) but the prediction accuracy is substantially deteriorated from 62% to 53%. Here we stress that the average advanced prediction time is 8.05 seconds under the aggressive approach, which is significantly larger than the values reported in the literature [10], [12], [16], [17], as shown in Fig. 9. Vehicle drivers have on average 8.05 seconds to take precaution, indicating a safer and more comfortable driving condition.

TABLE IV
PREDICTION RESULTS

| Metric | MLP model | Aggressive | Conservative |
| --- | --- | --- | --- |
| TPR | 0.62 | 0.75 | 0.58 |
| FPR | 0.32 | 0.46 | 0.31 |
| Prediction accuracy | 0.62 | 0.75 | 0.53 |
| Average advanced prediction time (s) | 6.35 | 8.05 | 7.44 |

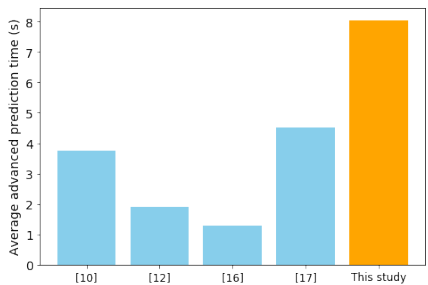

Fig. 9. Comparison of advanced prediction time

## VI. Conclusions and Future Work

This study built a long-term lane change maneuver prediction MLP model without using any lateral or angle information. Based on the real-world NGSIM traffic dataset, the time-window labeling scheme was extended by adding a time gap $\tau_g$ between positive samples and negative samples, and the performance of different values of $\tau_g$ was evaluated. In the run-time evaluation, an aggressive approach and a conservative approach were proposed to smooth predictions from the model. The prediction model is able to capture 75% of real lane change maneuvers with an average advanced prediction time of 8.05 seconds, which significantly outperforms other existing models.

There are several extensions that can be done to further improve the prediction of lane change maneuvers. First, a probabilistic approach can be developed to output a probabilistic prediction and capture the uncertainty in predictions. Second, an RNN based prediction model which captures the time series characteristics can be further refined and tested if more data is collected.

## Acknowledgment

The contents of this paper only reflect the views of the authors, who are responsible for the facts and the accuracy of the data presented herein. The contents do not necessarily reflect the official views of Toyota Motor North America.